\documentclass[letterpaper, 10 pt, conference]{ieeeconf}
\usepackage{amsmath}
\usepackage{amsfonts}
\usepackage{amssymb}
\usepackage{graphicx}
\usepackage{pgfplots}
\usepackage{hyperref}
\usepackage{pdfpages}
\usepackage{multirow}
\usepackage{todonotes}
\renewcommand\footnotemark{}

\renewcommand{\baselinestretch}{0.99}
\IEEEoverridecommandlockouts

\title{Multimodal Interaction and Intention Communication for Industrial Robots}

\author{Tim Schreiter$^{1*}$, Andrey Rudenko$^{2*}$, Jens V. Rüppel$^{1}$, Martin Magnusson$^{3}$, Achim J. Lilienthal$^{1,3}$
    \thanks{$^{1}$ \raggedright {Technical University of Munich, MIRMI,
    Chair of Perception for Intelligent Systems, Germany, \url{tim.schreiter@tum.de}}}
    \thanks{$^{2}$ Bosch Corporate Research, Germany,\url{andrey.rudenko@de.bosch.com}
    (${^*}$Shared first author)}     
    \thanks{$^{3}$ \"Orebro University, Sweden}
}

\date{}
\begin{document}

\maketitle

\begin{abstract}
Successful adoption of industrial robots will strongly depend on their ability to safely and efficiently operate in human environments, engage in natural communication, understand their users, and express intentions intuitively while avoiding unnecessary distractions. To achieve this advanced level of Human-Robot Interaction (HRI), robots need to acquire and incorporate knowledge of their users' tasks and environment and adopt multimodal communication approaches with expressive cues that combine speech, movement, gazes, and other modalities. This paper presents several methods to design, enhance, and evaluate expressive HRI systems for non-humanoid industrial robots. We present the concept of a small anthropomorphic robot communicating as a proxy for its non-humanoid host, such as a forklift. We developed a multimodal and LLM-enhanced communication framework for this robot and evaluated it in several lab experiments, using gaze tracking and motion capture to quantify how users perceive the robot and measure the task progress. 
\end{abstract}

\section{Introduction}
\label{sec:introduction}

Robots are increasingly used in shared environments with humans, making effective communication necessary for successful human-robot interaction. Many aspects of robot behavior define successful communication: generating clear, concise, and timely messages, supporting these messages with appropriate signals (verbal and non-verbal), directing the attention towards the relevant parts of the task and the environment while avoiding unnecessary distractions, and reading user feedback and task engagement from non-verbal cues such as position in space, gestures, and gaze direction. Combining these elements in a system that naturally fits dynamic human environments is challenging. Robots are often limited by their native design, making it difficult for them to produce legible social cues. Nevertheless, strong communication abilities are crucial in industrial contexts, where effective collaboration between humans and robots relies on mutual understanding and predictable behavior.

\begin{figure}
    \centering
    \includegraphics[width=\linewidth]{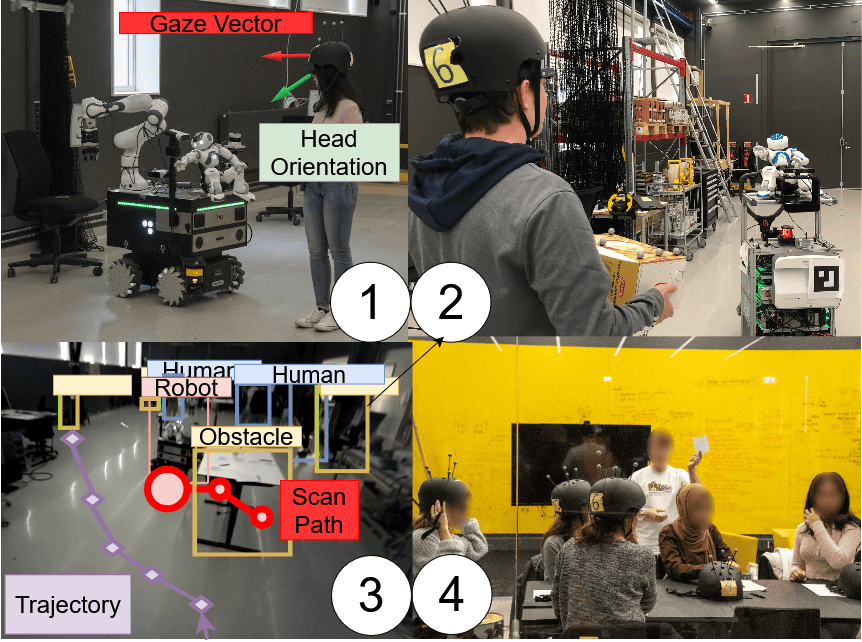}
    \caption{\textbf{Focus points and methods in our HRI Studies:}
(1) Anthropomorphic Communication Proxy for non-humanoid platforms
(2) Multimodal and LLM-enhanced communication
(3) Gaze tracking and motion capture
(4) Controlled user studies}
    \label{fig1}
\end{figure}

As part of the EU project DARKO\footnote{\url{https://darko-project.eu/}}, we develop methods for the next generation of agile production robots that are aware of humans and their intentions to smoothly and intuitively interact with them. Key to our research are \textit{Transferability} and \textit{Quantification} aspects of our methods. Aiming to address the inherent need to design transferable solutions to HRI that can be applied and verified on different robotic platforms \cite{cha2018survey}, we develop the concept of an ``Anthropomorphic Robotic Mock Driver'' (ARMoD) to communicate on behalf of the non-humanoid host platform. Here, we investigate the application of the ARMoD supporting the communication of an industrial robot in a representative interaction, involving approach, spoken instruction, and object manipulation (see Fig.~\ref{fig1}). To support the interaction in these settings, we utilize the developed expressive multimodal communication architecture, which includes robot speech, gaze, and gestures, directed to the task-relevant parts of the environment. To quantify the effect of the various communication styles, we adopt human gaze tracking as a measure of attention, intention, and task progress. Finally, we compare the traditional, partially scripted interaction to an LLM-enhanced one, investigating the potential to adapt the robot responses to the inherently dynamic and unpredictable human behavior.

\section{Methods}

\subsection{Lab studies}
\label{sec:lab}



We opted to investigate human-robot interaction scenarios in controlled laboratory settings. Although online studies offer scalability \cite{toris2013bringing} and ``in-the-wild'' experiments allow validation in complex social settings \cite{jung2018robots}, lab studies strike a balance for precise measurements of real human behavior and allow to isolate and condition the factors that may influence the interaction. Controlled environments also facilitate experimental repeatability and high-quality data collection. Specifically, in our recent studies \cite{schreiter2022effect,schreiter2023advantages,schreiter2024human,schreiter2025evaluating}, we investigated human-robot interaction in a scripted setup, which includes many elements found in industrial environments, and also more spontaneous interactions, which are recorded as part of a large-scope study of indoor human motion \cite{schreiter2024thor}.

The scripted interaction features several steps, relevant for industrial robots. Participants are instructed to deliver a tin can to the table, where they are approached by the robot asking for their assistance. The robot asks to pick up a large box and place it on the forks of the forklift. The interaction concludes afterwards with a disengagement.

In contrast, the spontaneous interactions \cite{schreiter2024thor} involve the robot being approached by a person in different positions and settings. The robot communicates its next goal point and asks the person to accompany it. In these interactions, the robot moves either differentially or omnidirectionally.

\subsection{Multimodal and LLM-enhanced communication}\label{sec:llms}


When actively interacting with the user, robots can benefit from a wide variety of available modalities \cite{pascher2023communicate} to support and enrich their messages with non-verbal cues, acknowledge the reception of user's commands and refer to the objects in the environment. Research shows that multimodal approaches can improve interaction speed, accuracy, and naturalness by better mimicking human communication patterns \cite{salem2011friendly, kompatsiari2019measuring}. Furthermore, users will likely evaluate robots capable of multimodal communication more positively \cite{salem2011friendly}.
Following the interaction designs presented in Sec. ~\ref{sec:lab}, we implemented a multimodal communication design to support users' tasks and compared it to verbal-only conditions.

In addition to expressive multimodal communication, robots need to flexibly adapt their messages and actions to the environment's context of the environment and the status of the interaction. We use Large Language Models (LLMs), owing to their advanced reasoning capabilities, to extend our multimodal communication framework with real-time context interpretation and natural language response generation capabilities. The potential to improve the interaction flow, in comparison to more traditional pre-scripted behavior, is yet to be qualified in practice. Specifically, we compared the scripted interaction from Sec.~\ref{sec:lab} with an equivalent one, which benefits from LLM-enhanced responses \cite{schreiter2025evaluating}.


\subsection{Anthropomorphic communication proxies}
\label{sec:armod}

To be capable of expressive multimodal communication, robots need specialized modalities that are intuitively interpretable by people. However, the function-driven design of non-humanoid service and industrial robots limits their ability to express human-readable cues. To address these conflicting requirements, we introduced the Anthropomorphic Robotic Mock Driver (ARMoD) concept of a small robotic entity that extends the host system (e.g., a non-humanoid robot) and can communicate with natural, human-readable signals. ARMoD is designed to standardize communication patterns across diverse robotic platforms \cite{schreiter2023advantages}. We designed a multimodal communication protocol for ARMoD that combines speech, gaze, and referential gestures. The ARMoD was deployed in all interactions, presented in Sec.~\ref{sec:lab}

\begin{figure}
\vspace{0.2cm}
    \centering
    \includegraphics[width=\linewidth]{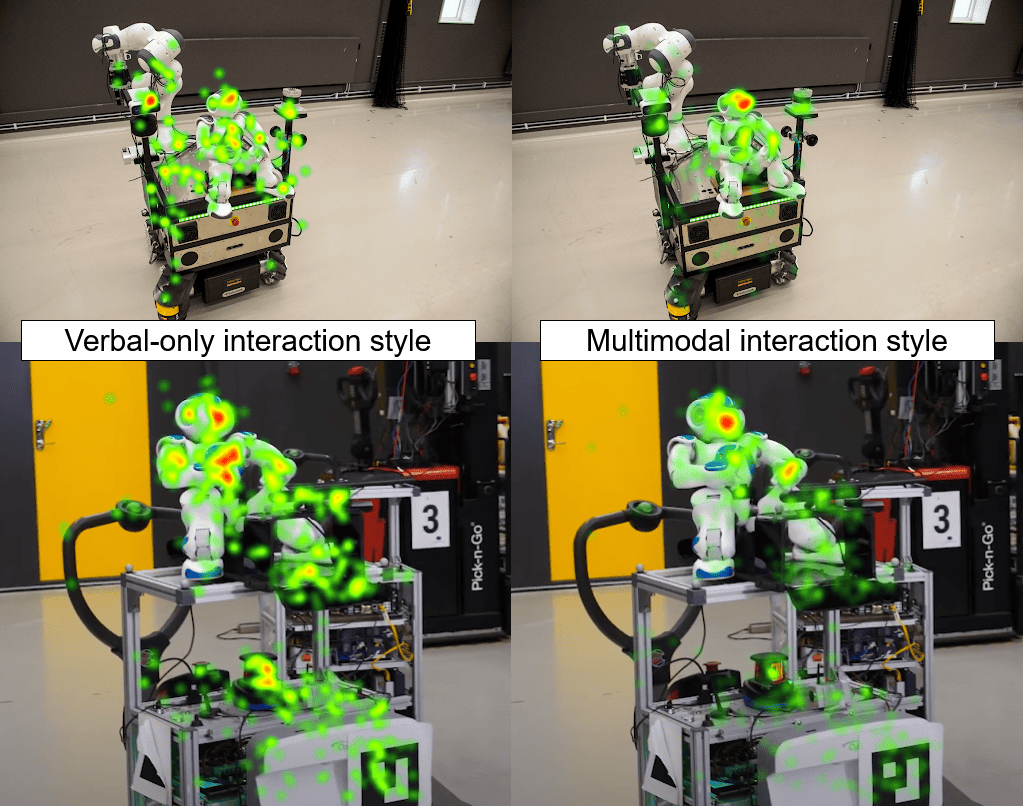}
    \caption{Heatmaps showing participant gaze distribution on two robot platforms (including the ARMoD) for two interaction styles (verbal-only and multimodal).
    In the multimodal style, eye fixations are more concentrated on the ARMoD humanoid robot. 
    }
    \label{fig2}
\end{figure}



\subsection{Gaze tracking and motion capture}
\label{sec:thor}

One of the key challenges of HRI research is assessing the benefit of novel robot behaviors \cite{cha2018survey}. In our experiments, we use gaze tracking to objectively measure user eye movements and head rotations \cite{schreiter2024human} as participants navigate, explore, and manipulate objects in shared environments with robots. By integrating these measures with motion capture data, we directly correlate gaze behavior with task performance and attention distribution during motion, providing critical insights into how users perceive and adjust their behavior in the presence of our robot communication strategies \cite{schreiter2025evaluating}. This approach validates the effectiveness of our novel behaviors and informs the improvement of HRI systems for more natural and intuitive interactions.

\section{Results}

Combining the insights from 3D position and head orientation motion capture and gaze tracking, we could derive several notable conclusions about human behavior in scripted and spontaneous interactions with robots.

We found that users react faster in collaborative tasks with the robot equipped with an ARMoD and multimodal interaction style. When the robot gives instructions supported by gaze, users are quicker to localize goal points and objects of interest \cite{schreiter2023advantages}. We also notice the concentration of fixations on the ARMoD using a multimodal communication style, as opposed to a verbal-only interaction (see Fig.~\ref{fig2}). In contrast, we do not find significant differences between the perception of the robot moving differentially vs. omnidirectionally in the THÖR-MAGNI dataset \cite{schreiter2024human}. Similarly, participants did not achieve higher task efficiency when the robot guided the interaction with LLM-enhanced responses, as compared to the fully scripted scenario \cite{schreiter2025evaluating}.

These examples illustrate our attempts to achieve more efficient and natural human-robot interaction, that can be transferred to different robots and quantified beyond subjective questionnaire ratings. In our future work, we aim to transfer these communication and evaluation methods to other domains outside of industry, such as elderly care.


\bibliographystyle{IEEEtran}
\bibliography{prediction_state_of_the_art}

\end{document}